\documentclass[10pt,conference]{IEEEtran}
\usepackage{amsmath} 
\usepackage{algorithm}  
\usepackage{algpseudocode,algorithmicx}
\usepackage{slashed} 
\usepackage{wasysym}
\usepackage{bm}
\usepackage{mathrsfs}
\usepackage{graphicx}
\usepackage{subfigure}

\usepackage{cite}
\ifCLASSINFOpdf
\else
\fi
\usepackage{amsmath}
\usepackage{array}
\usepackage{stfloats}

\begin{document}

\title{Doppler-Radar Based Hand Gesture Recognition System Using Convolutional Neural Networks}


\author{Jiajun Zhang, Jinkun Tao, Jiangtao Huangfu, Zhiguo Shi\\
	\IEEEauthorblockA{
		College of Information Science \& Electronic Engineering\\
		Zhejiang University, Hangzhou, China\\
		{\{justinzhang,~jktao,~huangfujt,~shizg\}@zju.edu.cn}
	}\\
}

\maketitle

\begin{abstract}
Hand gesture recognition has long been a study topic in the field of Human Computer Interaction. Traditional camera-based hand gesture recognition systems can not work properly under dark circumstances. In this paper, a Doppler-Radar based hand gesture recognition system using convolutional neural networks is proposed. A cost-effective Doppler radar sensor with dual receiving channels at 5.8GHz is used to acquire a big database of four standard gestures. The received hand gesture signals are then processed with time-frequency analysis. Convolutional neural networks are used to classify different gestures. Experimental results verify the effectiveness of the system with an accuracy of 98\%. Besides, related factors such as recognition distance and gesture scale are investigated.
\end{abstract}

\IEEEpeerreviewmaketitle

\section{Introduction}
Hand gesture recognition has been a long been a study topic in the field of computer science. It has been regarded as a way of interactions between machine and human. This technology enables computers to understand human instructions without traditional interaction hardware like the mouse and keyboard. Traditional hand gesture recognition systems are based primarily on cameras and image processing algorithms\cite{hjelmaas2001face}. While camera-based hand gesture recognition system provides reliable recognition rate, they have limitations. The most obvious one is that it is highly impacted by the brightness of light\cite{zhao2003face}. In addition, the high demand of computational and power resources also constrain them from being adopted when there are limited resources of processors and batteries\cite{gavrila1999visual}. Besides, the nature of camera-based recognition system will cause the privacy concern in public use.

Recently, hand gesture recognition based on radar has begun to gain interests in public\cite{molchanov2015multi,molchanov2015short,arbabian201394}. Compare with traditional methods, radar-based hand gesture recognition has its own merits. First of all, while camera can hardly capture a clear image under dim light, radar signal is not affected and can be widely used under dark situations. Secondly, continuous wave Doppler radar sensors detecting the Doppler effect produced by moving objects scattered by radio frequency (RF) signals can be implemented with cost-effective architecture. That is, the frequency of the Doppler phase caused by human gestures is only shifted by a limit of a few hertz, and the price of ADC converters and baseband devices are cost effective. Therefore, the radar-based hand gesture system has obvious advantages for practical use. 

However, compared to the massive number of researches on the camera-based hand gesture recognition system, there are few literatures to use radar for gesture recognition until recently. Some of the studies focus on radar-based hand gesture recognition on Band E or WLAN\cite{hugler2016rcs}\cite{pu2013whole}. In \cite{hugler2016rcs}, researchers use the mono-static radar cross section measurements of a human hand for radar-based gesture recognition at E-band, whose frequency range from 60G Hz to 90G Hz But apparently, E-band is too high and too expensive for massive use in real life. In \cite{pu2013whole}, they used Wi-Fi signal to do sensing or recognize gestures of human in a home area. Because Wi-Fi signals can travel through walls, this system indeed enables whole home gesture recognition using fewer wireless sources. However, with so many routers using Wi-Fi technology in daily life, 2.4GHz Wi-Fi seems too crowded for massive use. 

On the other hand, there are several works on human falling detection, which is a kind of body gesture, by using microwave radar in recent years. Some representative works are presented in \cite{mercuri2013analysis,peng2016fmcw,zhou2016ultra}. High accuracy falling detection from normal movements was achieved by Zig-bee module in computer\cite{mercuri2013analysis}. Besides, in \cite{peng2016fmcw}, a coherent frequency-modulated continuous-wave (FMCW) radar sensor is designed and tested for long-term wireless fall detection in home and hospitals. By analyzing the radar cross section (RCS), range, and Doppler changes in the Inverse-Synthetic-Aperture-Radar (ISAR) image during the subject's movement, falling can be distinguished from normal movements such as sitting. Moreover, some researchers adopt both video and ultra-band radar in their attempts\cite{zhou2016ultra}. They applied hidden Markov models to train the features extracted from received signals to discern the types of motion. However, the difference between hand gesture recognition and fall detection lies in that, hand gesture recognition detection needs more fine-grained signal processing.

In this paper, we propose a hand gesture system based on Doppler-Radar using convolutional neural networks (CNN). Unlike many researches focusing on modeling received radar signal of different hand gestures, our approach focuses on establishing relationship between radar signal and hand gestures based on a sample database. Specifically, we adopt a Doppler radar with double signal channels at 5.8GHz to acquire a big data sample of four standard common-use gestures; Then we apply short-time Fourier transform and continuous wavelet transform as two major time-frequency analysis methods to the received signal; At last, with the results of time-frequency analysis, we use convolutional neural networks, a machine learning algorithm to perform the classification. Moreover, we discuss the effect of the following two factors on the accuracy of gesture recognition: distances between the gesture and sensor and scale of gesture. The results show that, accuracy may drop a little bit but still keeps on a high rate when the distance between gestures and sensor becomes larger. And the scale of the gesture does not vary much about the accuracy rate. However, the nature of the convolution neural networks implies that more samples are needed to achieve a higher accuracy of different people. Our results show that the proposed hand gesture recognition system based on Doppler-Radar and CNN is able to have a great specific gesture recognized in an extraordinary accurate way.

The whole paper is structured in the following way. The second section describes the problem and introduces hardware structures and implementation of the system. Section III introduces the system architecture, including three parts: data acquisition, time-frequency analysis and classification, and discusses the results of our experiments. Finally, conclusion remarks are drawn in Section IV.

\section{Hardware Architecture and Implementation}

\subsection{Problem Description}
Intuitively, hand gesture recognition falls into the category of pattern recognition. The most frequent pattern recognition of hand gesture is based on visions, images or videos, like mentioned before. Since each hand gesture has its particular features, the patterns of every hand gesture can be just extracted and processed into easy categorical features. Finally, through machine learning mechanisms, using the processed features as training data, a classification model can be got. It can classify testing hand gestures according to their patterns.

The main procession of hand gesture recognition based on 5.8GHz Doppler radar is just like the one with images or videos. The patterns of visions are extracted directly from the taken image. But the patterns of radar gesture are extracted from the scattered waveforms. Meanwhile, the signal wave will not only be echo reflected by the hand and any skins, but also any surrounding body part, especially when there exist complex background noises. Thus, the received signal could be a difficult puzzle for anyone who tried to solve the signal due to numerous micro-motions. Therefore, extracted features would be slightly different from the actual ones. 

Therefore, the main point is to look for features from different perspective and find those features which are suitable for scattered radar waveforms and cannot be severely influenced by background noise. Then, we train the data with machine learning algorithms. In this paper, we analyze the data from time and frequency perspectives and get the features. And then we use convolution neural network to train the data and get the classification model.

In our case, our goal is to differentiate four gestures, which are hand gestures drawing a circle, a square, a tick and a cross. The following experiments are all based on these four gestures.

\subsection{Hardware Architecture}
\begin{figure}[htb]
	\centerline{\includegraphics[height=5cm,clip]{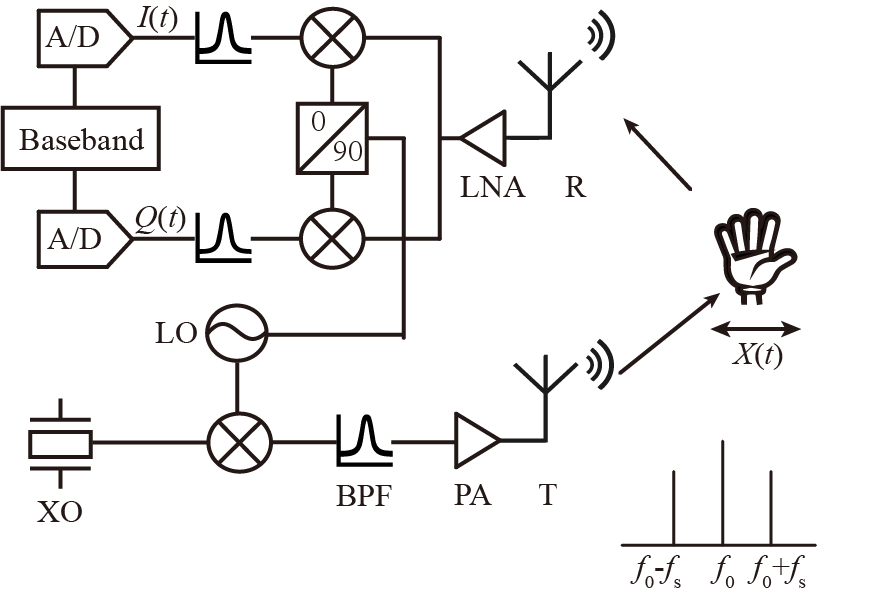}}
	\caption{Hardware prototype}
	\label{hardware}		
\end{figure}

To have a low-cost and high-quality raw signal, we designed an architecture with double sub-carrier modulation centered at 5.8GHz band IF architecture\cite{fan2016wireless}. And you can see the whole prototype in \ref{hardware}.

Sub-carrier modulation is commonly seen in motion detection, including contacts of several channels. The baseband signal will be modulated into several sub-carriers. And then it will transformed to Radio Frequency carrier. Let us see how it works in Fig. \ref{hardware}. A crystal oscillator will generate desired frequency at the top left of the picture. And it is with $f_{sub}$ deep lower frequency. Then the signal will do the local signal mixing. At last, the mixer unit passes through a bandpass filter. The the signal then will be power amplified before transmitting through TX antenna. The transmitted signal now has two sidebands at the local frequency. Transmitted sub-carriers along with the local oscillation are not only reflected from the hand gesture, but also from the interference noise. Thus the reflected signal is down-sampled by local frequency signal, getting a coherent, quadrature zero-IF conversion. Then the signal will go through a bandpass filter. It is used to take away the direct current and also some noise from the background of body. The ADCs here are used to transform the analog signal to digital signal. And the sub-carrier will be down-converted. 
We use super narrow-band ADCs to improve the quality and meanwhile the cost of the architecture.  We should know that after processing the downside by mixer, the bandwidth cannot be transformed out of the low range.

\begin{figure}[htb]
	\centerline{\includegraphics[height=4cm,clip]{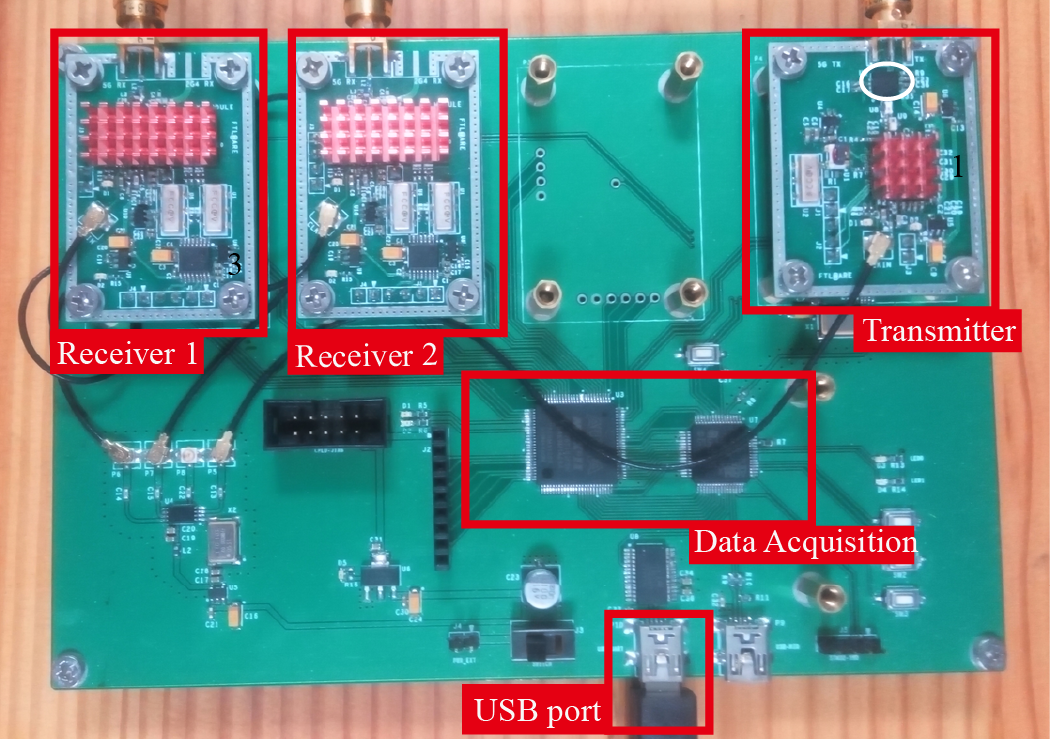}}
	\caption{Hardware structure}
	\label{hardwarereal}		
\end{figure}

Fig. \ref{hardwarereal} is the finished version of system. The system is designed to work on band of 5.8GHz. First of all, there are typical double channels to receive the channels. These chips are moderated from the some WI-FI products. Since we have to consider the cost and the price of the chips, we chose these due to its cheap price and relatively better quality. However, the limitation of the chips is that they can only be settled at 6MHZ. The two Analog to Digital converters are used in down-converters The two sideband are between 100Hz. We use RS232 to transmit the data to collection software on PC.

\section{System Design and Analysis}

The operation flow of hand gesture recognition system is approved as Fig. \ref{outline}. The first step is data acquisition. The database is essentially composed of the received signal samples from different hand gesture. We establish four standard gestures (circle, square, tick, cross) as examples in our research. Since the whole system is based on the process of a big gesture sample database, we have to build a large hand gesture database with the same background but different parameters, such as the distance between the gesture and the hardware, the scale of the gesture, different test hands. 

After developing the huge hand gesture signal database, we do time-frequency analysis of the time-domain signals from the database. Here, we use two stable and efficient time-frequency analysis algorithms. The first is known as short-time Fourier transform and the second is called continuous wavelet transform. Both algorithms give the perspective of how the frequency of the back scattered signal is changed during gesture moving and can be used as an indication of the difference between different gestures. 

Finally, we utilize the results of time-frequency analysis as features in classification step. Here we use convolutional neural network (CNN) as classification algorithm, which is a classic method of machine learning. We randomly choose half of the database samples as training data and the other half as testing data. Following are the specific design of each processing step in the system:

\begin{figure}[htb]
	\centerline{\includegraphics[height=5cm,clip]{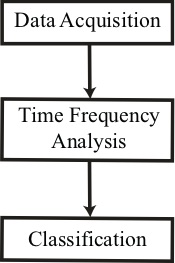}}
	\caption{System Operation Flow}
	\label{outline} 		
	\end{figure}

\subsection{Data Acquisition}
\begin{figure}[htb]
	\centerline{\includegraphics[height=4cm,clip]{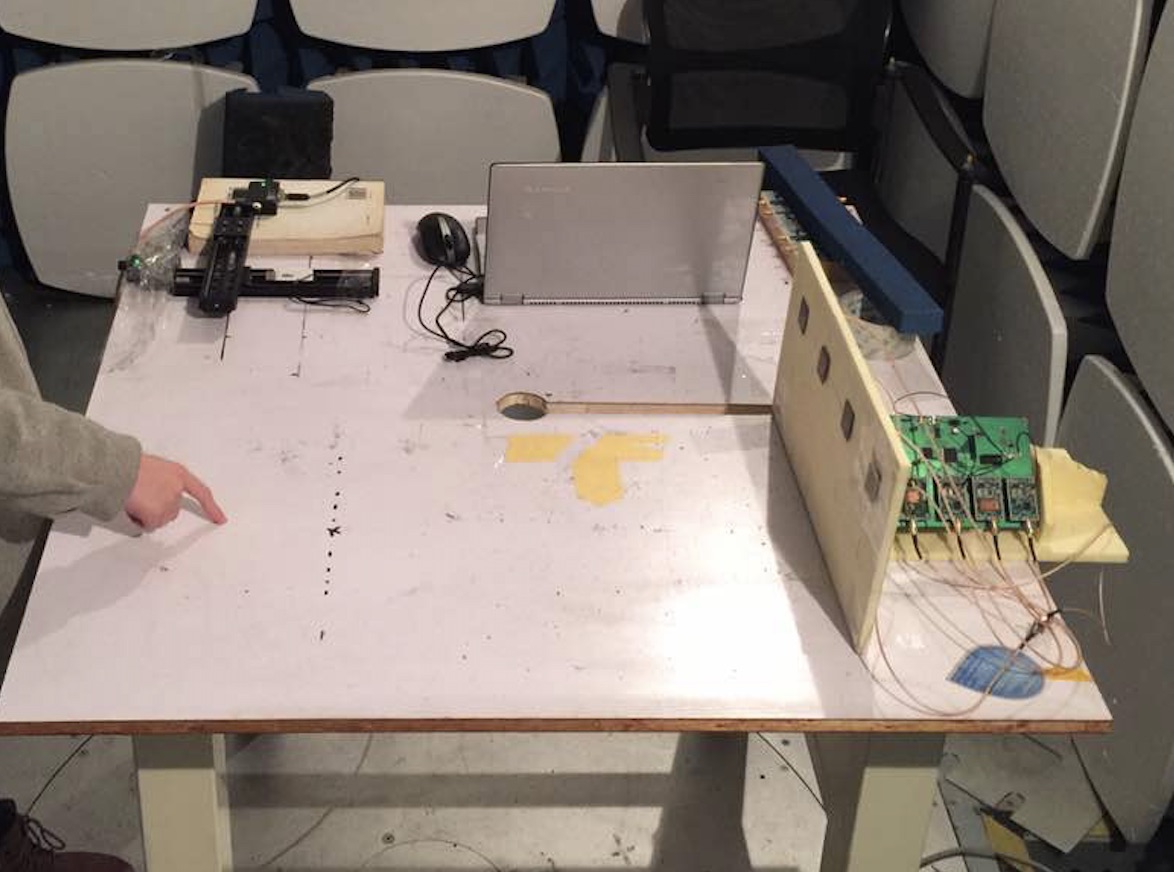}}
	\caption{Experiment setup for data acquisition}
	\label{posing} 		
\end{figure}

Fig. \ref{posing} shows the experiment setup of the data acquisition process. Three regular circularly antennas are uniformly placed on a foam structure with 10cm space in between, acting as R1 (Left), T (middle) and R2 (right). Since the method is based on big data, we need to obtain a larger database of hand gestures. And in order to consider what affect accuracy of recognition rate of gesture, we acquired hand gesture samples in different situations.  This polarizing setting can ensure scattering from different shapes of objects can be identified by the receiving antennas. In the following experiments, we do the experiments in a sealed room with less noise and we reduce other effects by reducing the amount of people doing the experiments.
Although doing that, noises from the body movement can also be interfere the result of the received signal. The problem can be solved by tunning the transmitting power so that the receiver only detects the movements of the hand.

\begin{figure}[htb]
	\centering
	\subfigure[]{
		\includegraphics[height=3cm,width=4cm]{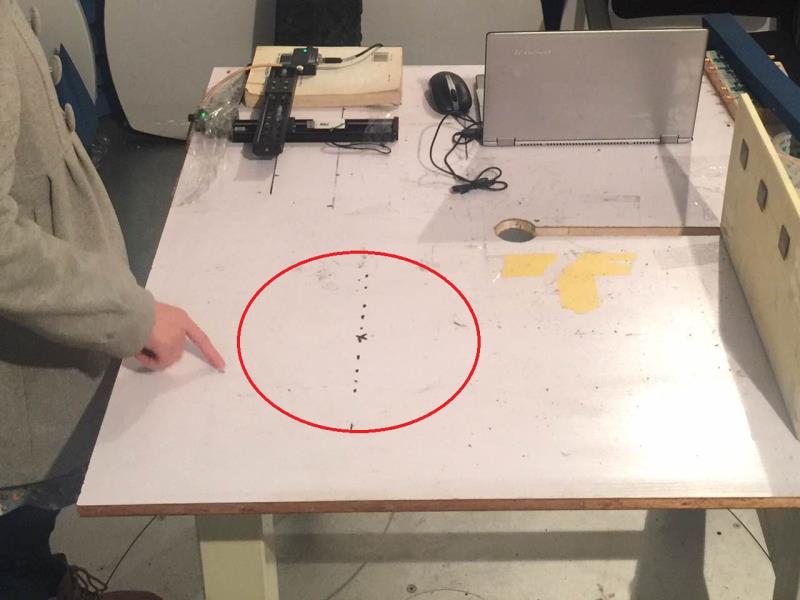}}
	\hspace{0.1in}
	\subfigure[]{
		\includegraphics[height=3cm,width=4cm]{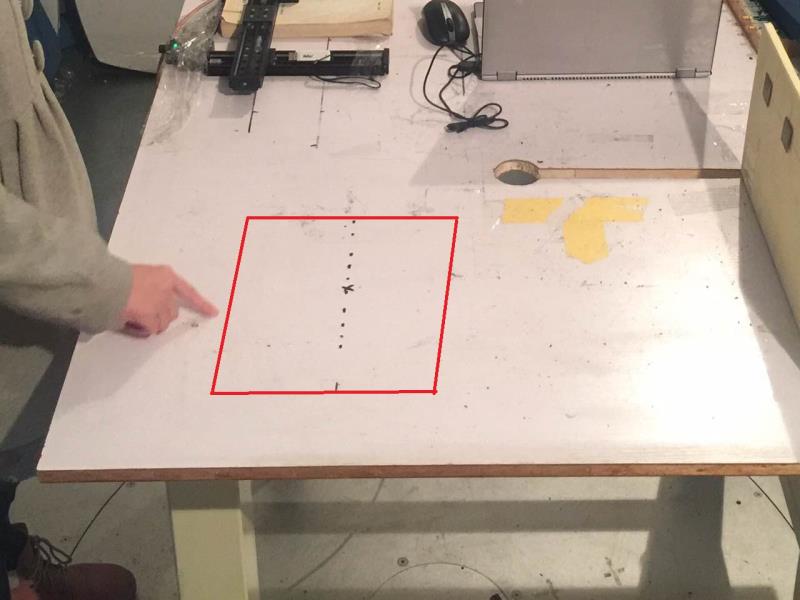}}
	\hspace{0.1in}	
	\subfigure[]{
		\includegraphics[height=3cm,width=4cm]{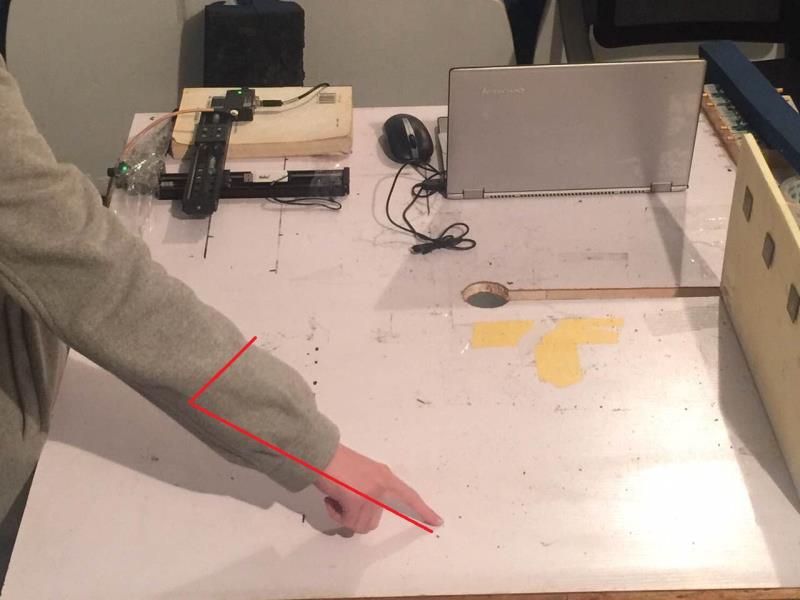}}	
	\hspace{0.1in}
	\subfigure[]{
		\includegraphics[height=3cm,width=4cm]{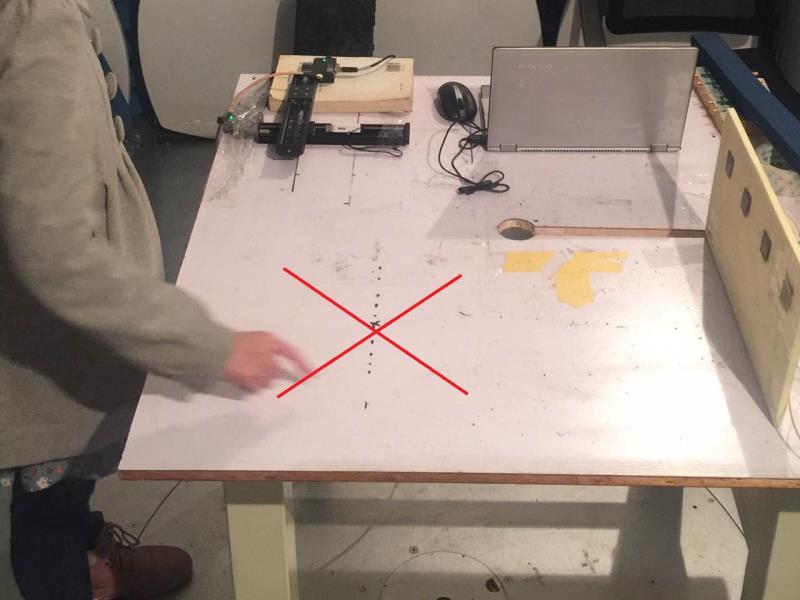}}	
	\caption{Four standard gestures:(a) Circle  (b) Square (c) Tick (d) Cross}
    \label{handgesture}
\end{figure}

In our hand gesture recognition system, we set four hand gestures as our standard hand gestures, which are circle, square, right, cross, as in Fig. \ref{handgesture}. All the gestures are captured at a same speed and the same time window of 1 second. And the gestures are in the horizontal plane in front of the antennas, as in Fig. \ref{posing}.

Since the method is based on big data, we need to obtain a larger database of hand gestures. And in order to consider what affect accuracy of recognition rate of gesture, we acquired hand gesture samples in different situations. It is true that there are many factors which may affect the accuracy of the recognition rate. To simplify the experiments, we only focus on two factors at this time: distance, scale of a gesture.

At last, a total of 9600 samples were acquired at different distances from the antennas, $d=[0.1,0.2,0.5]$ (in meters). For each distance, hand gestures were posing with two different scales, $r=[0.2,0.5]$ (in meters). In our experiments, for each gestures of certain distance and scale, we capture 50 samples of 4 volunteers each. In total, 9600 hand gesture samples were collected in the database.

\begin{figure}[htb]
	\centering
		\subfigure[]{
			\includegraphics[height=3cm,width=4cm]{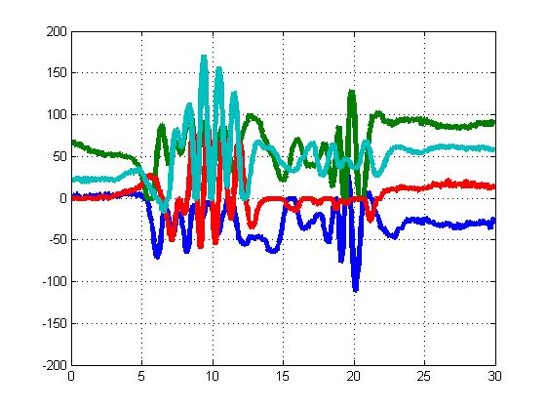}}
		\hspace{0.1in}
		\subfigure[]{
			\includegraphics[height=3cm,width=4cm]{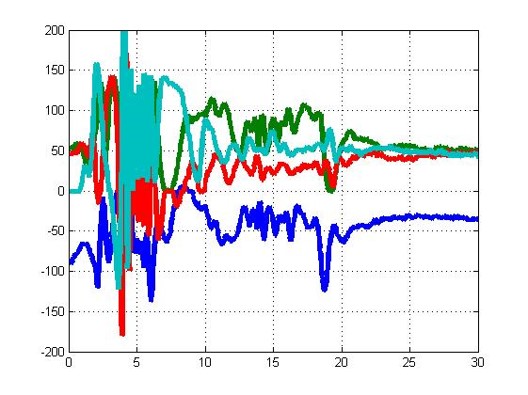}}
		\hspace{0.1in}
	    \subfigure[]{
	        \includegraphics[height=3cm,width=4cm]{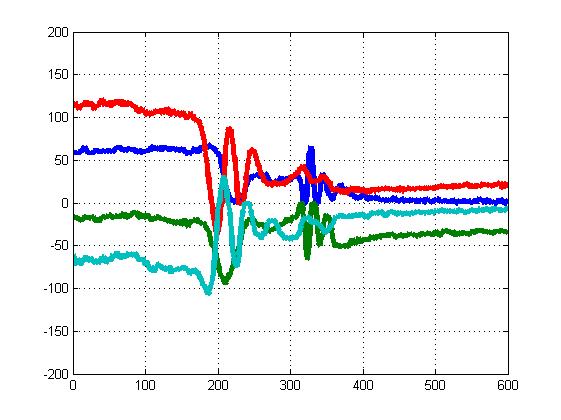}}
        \hspace{0.1in}
        \subfigure[]{
            \includegraphics[height=3cm,width=4cm]{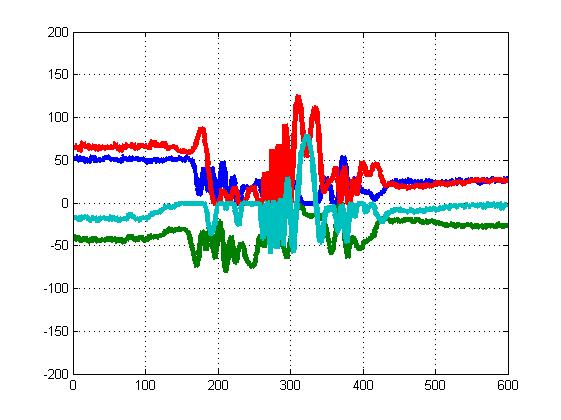}}
	\caption{Receiving signals in Time Domain. (a) Circle  (b) Square (c) Tick (d) Cross }
	\label{time}		
\end{figure}

Utilizing the hardware in the second section, we can record data using RS232 connecting to the USB port on the hardware. Then, we can have two channel signals continuously flowing from two receivers. We all know from the textbook that a quadrature receiver have two channels. The baseband output has two channels $I$ channel and $Q$ channel. In total, we have four channel signals. Fig. \ref{time} shows the signals in the time domain corresponding to four standard gestures.

As Fig. \ref{time} shows, the signals of the four gestures have a very big difference in the time domain. All the gestures have very large fluctuations in the early stages of time, but the fluctuation of square gesture is relatively greater. In addition, there are different jitters in the lasting seconds of the time domain between the circle gesture and the square gesture. As for the tick gesture, the fluctuation is bit small which may caused by its smaller movement of the gesture. And for the cross gesture, there is a peak that reflects its severe motion. There may exist some differences between the different samples. We can later exclude this effect through statistical methods.

\subsection{Time-Frequency Analysis}

In this subsection ,we apply time-frequency analysis to see more feature difference between four gestures. Time-frequency analysis is a two-dimensional figure describing the signal's spectrum change with time. The most common methods of time-frequency analysis are short-time Fourier transform (STFT)\cite{sejdic2009time} and continuous wavelet transform (CWT)\cite{mallat1999wavelet}.

The short-time Fourier transform, is a way to show signal in a more clear way. We can see from the STFT that how the signal's frequency changed through time. Compared to Fourier transform, it focuses on the instantaneous frequency information. In fact, the process of calculating STFTs is to cut a longer time signal into several time leg of same length. After that, we calculate the FFT on each leg of time individually. The spectrogram is computed as following equation:

\begin{align}
& STFT(t,\omega)=\int s(t')\omega(t'-t)e^{-j\omega t'}dt', \label{equ1}\\
& STFT(t,\omega)=\frac{1}{2\pi}e^{-j\omega t}\int s(\omega')W(\omega'-\omega)e^{j\omega' t}dw' \label{equ2}.
\end{align}

\begin{figure}[tb]
	\centering
	\subfigure[]{
		\includegraphics[height=3.3cm,width=4cm]{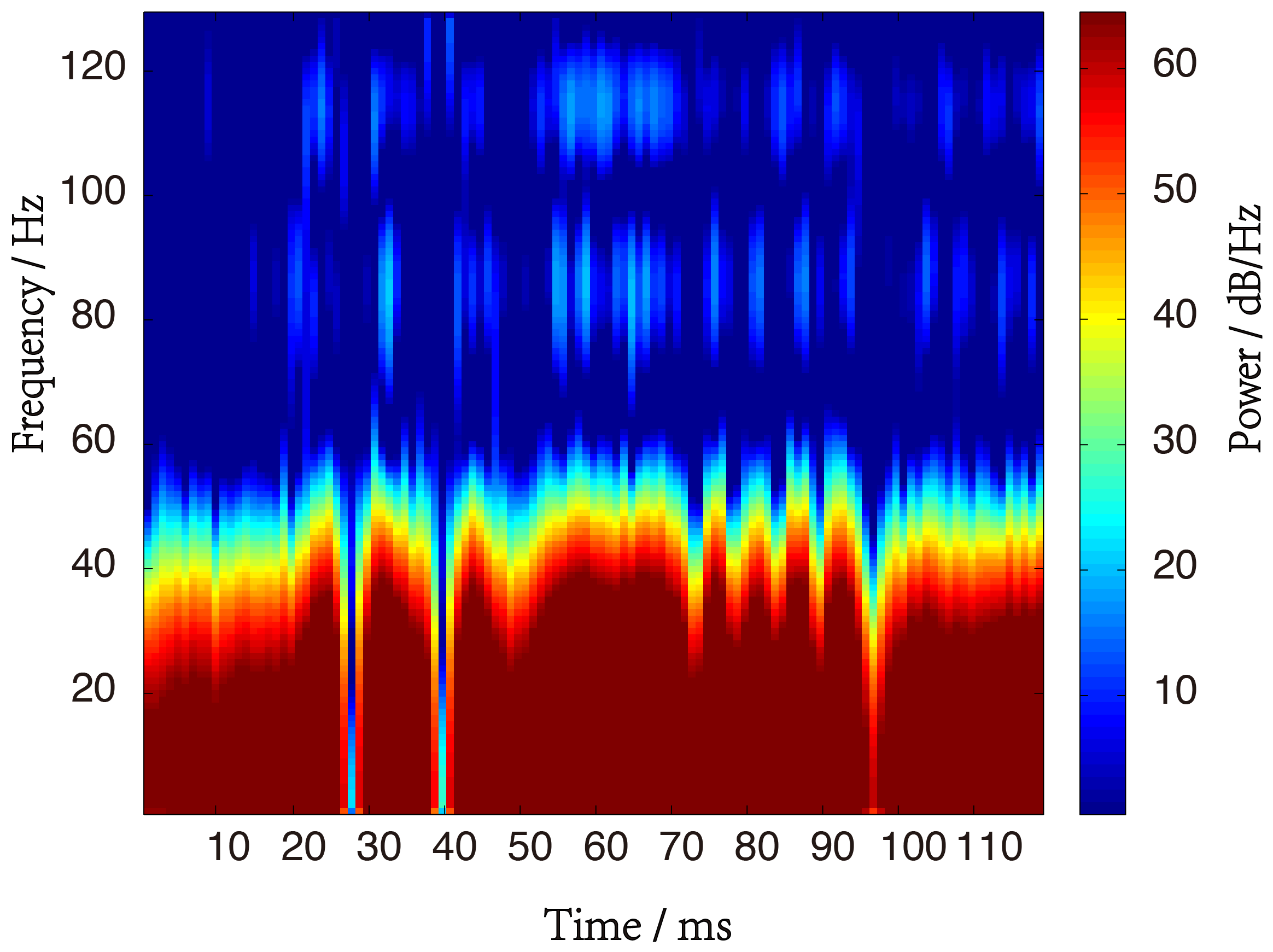}}
	\hspace{0.01in}
	\subfigure[]{
		\includegraphics[height=3.3cm,width=4cm]{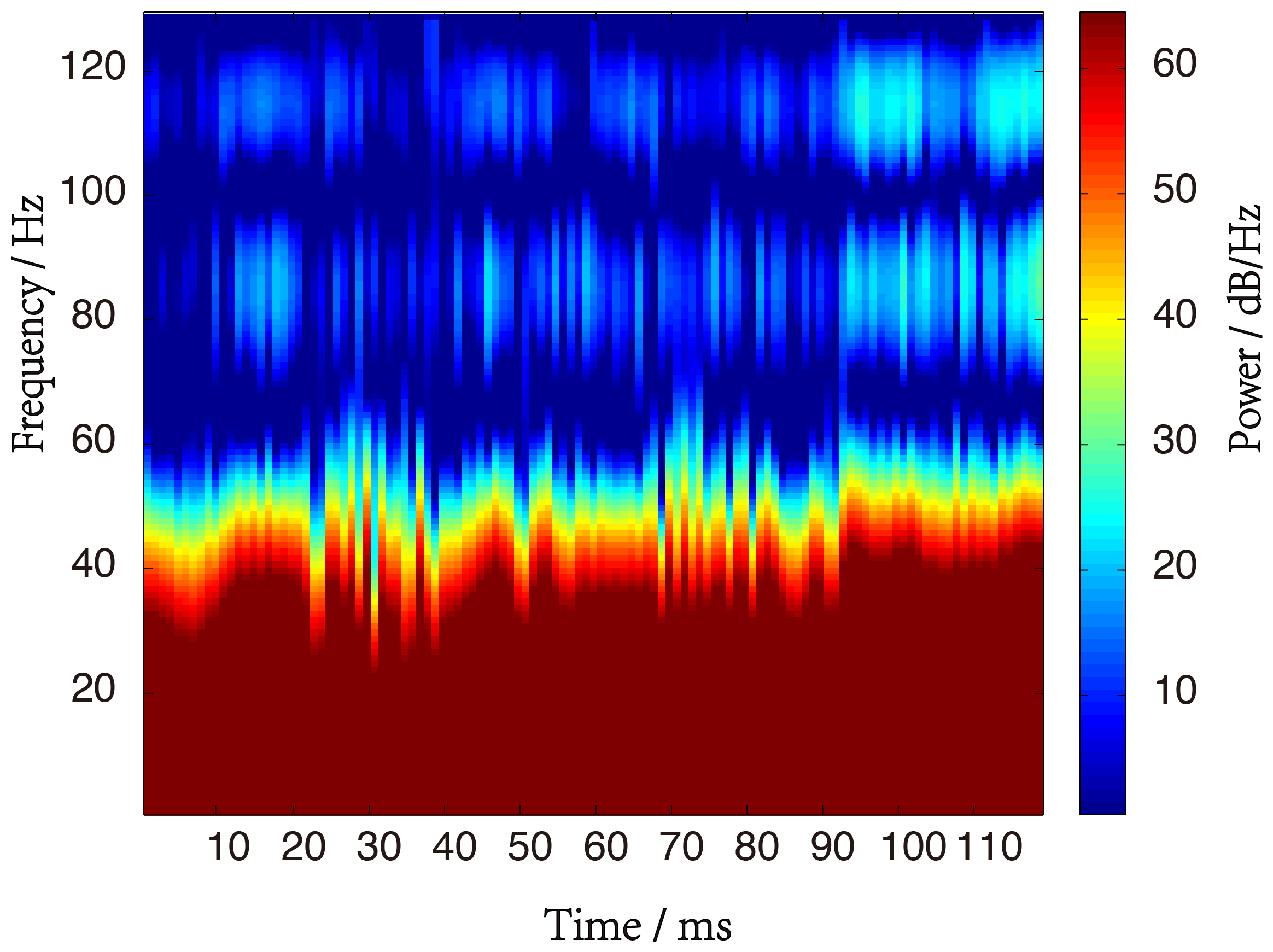}}	
	\subfigure[]{
		\includegraphics[height=3.3cm,width=4cm]{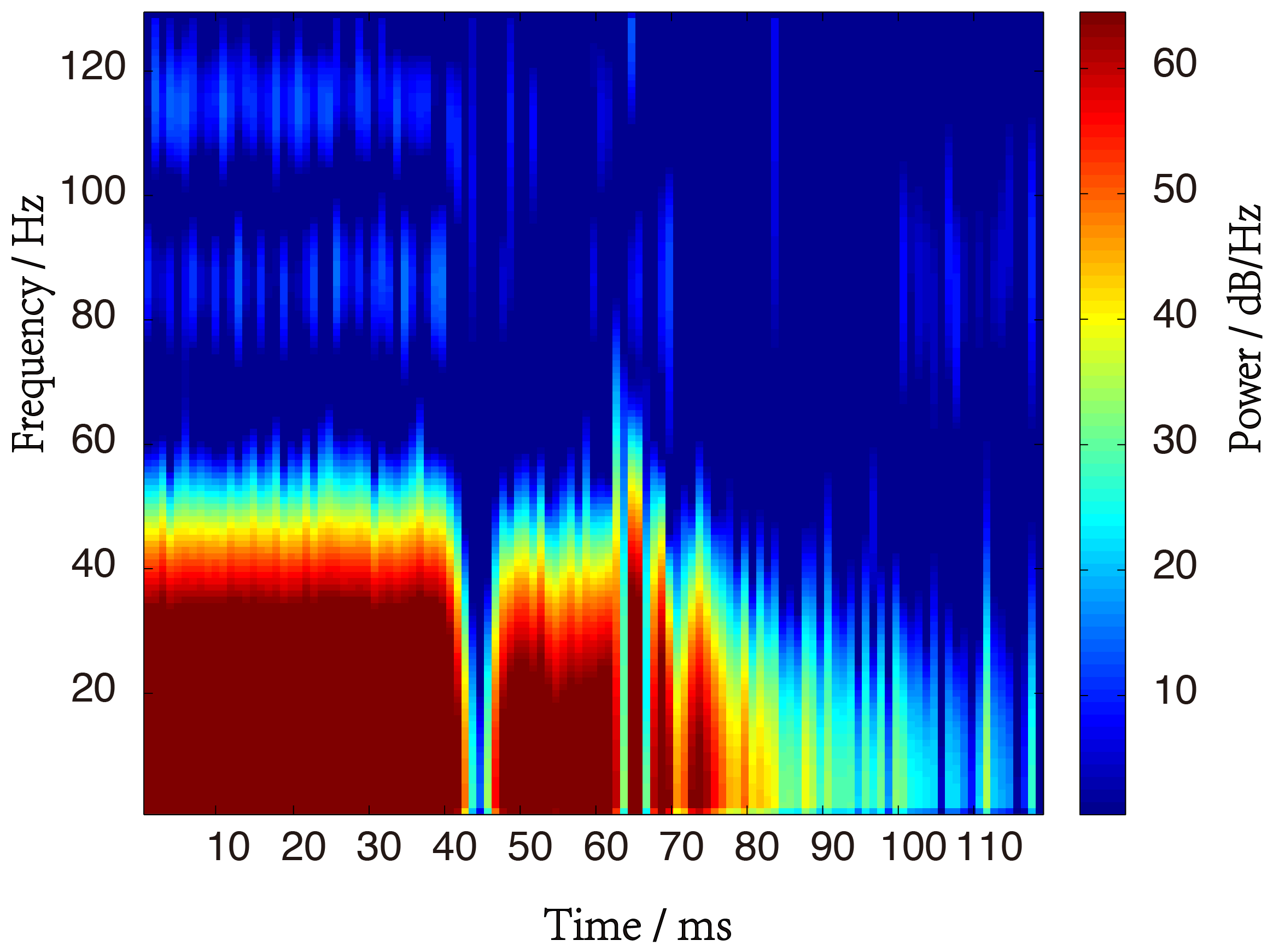}}
	\hspace{0.01in}
	\subfigure[]{
		\includegraphics[height=3.3cm,width=4cm]{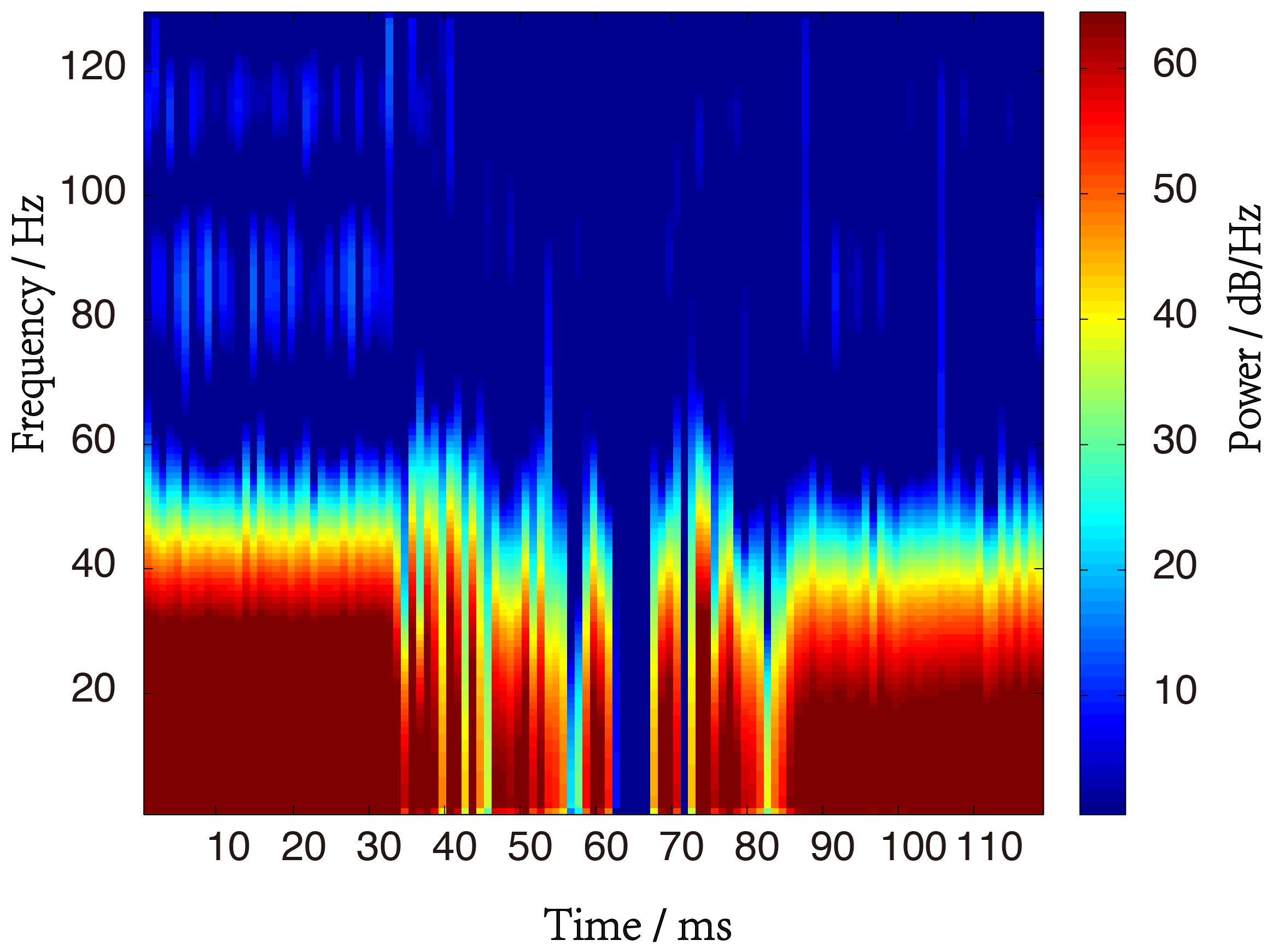}}	
	\caption{Time-frequency analysis using STFT. (a) Circle  (b) Square (c) Right (d) Cross}
	\label{STFT}		
\end{figure}

Fig. \ref{STFT} presents four time-frequency spectrograms using STFT method. The width of the window is 1 sec, and the moving range is 0.1 sec at a time. As we can see, the difference is quite noticeable. The spectrogram of square hand gesture is more evenly distributed than the circle hand gesture. As for the square gesture, the frequency is quite large at the beginning of the time but drops at 0.4s approximately. And there is always peaked in the middle of the test time in the outcome of the cross gesture.

However, the STFT has one disadvantage that the width of the time window is inversely proportional to the width of the frequency window. That means, the high resolution rate in the time domain will cause low resolution is frequency domain and vice versa. The width of sliding window restrains the frequency resolution rate of STFT. And CWT is an alternative way to solve this problem.

\begin{figure}[htb]
	\centering
	\subfigure[]{
		\includegraphics[height=3.3cm,width=4cm]{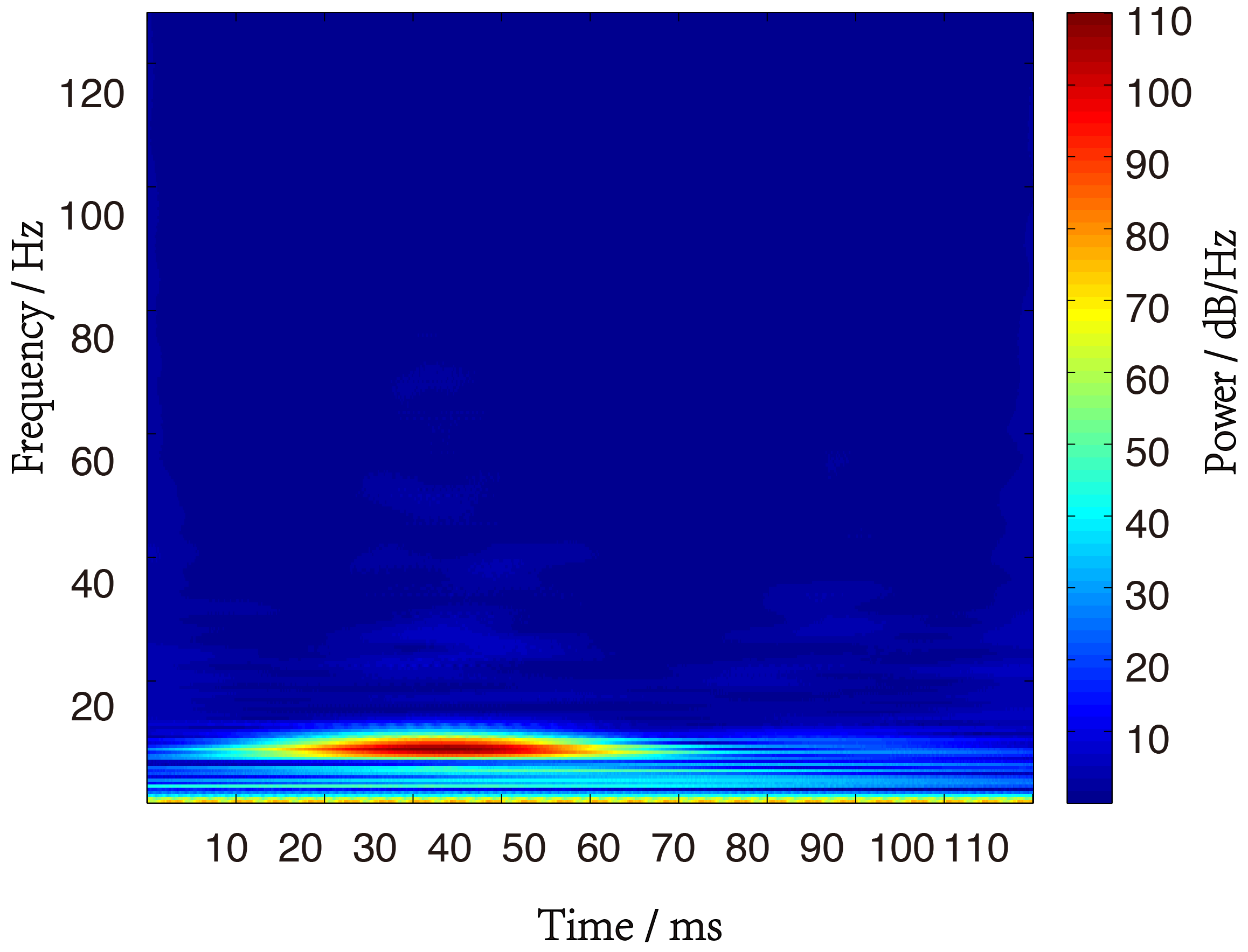}}
	\hspace{0.01in}
	\subfigure[]{
		\includegraphics[height=3.3cm,width=4cm]{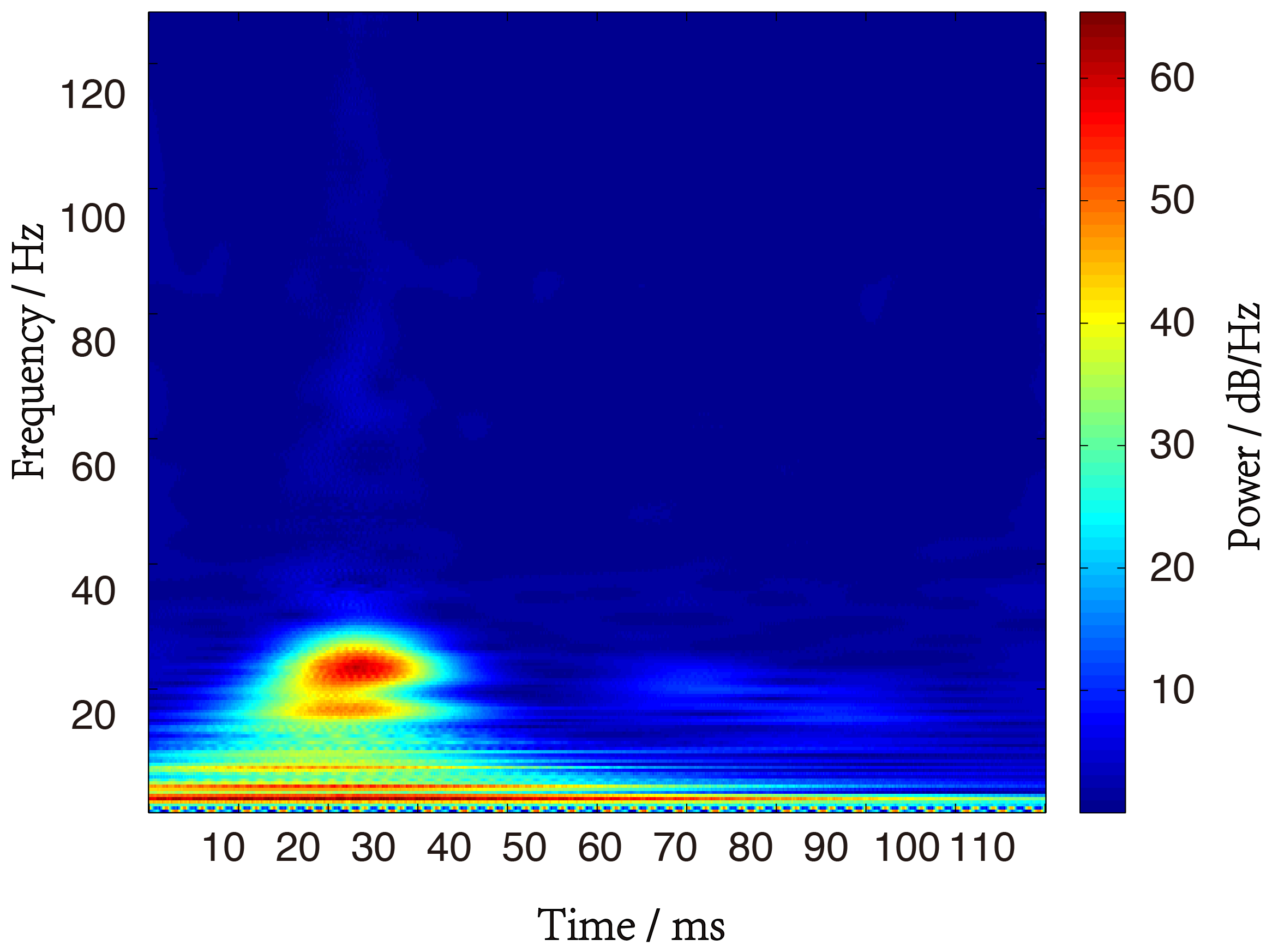}}	
	\subfigure[]{
		\includegraphics[height=3.3cm,width=4cm]{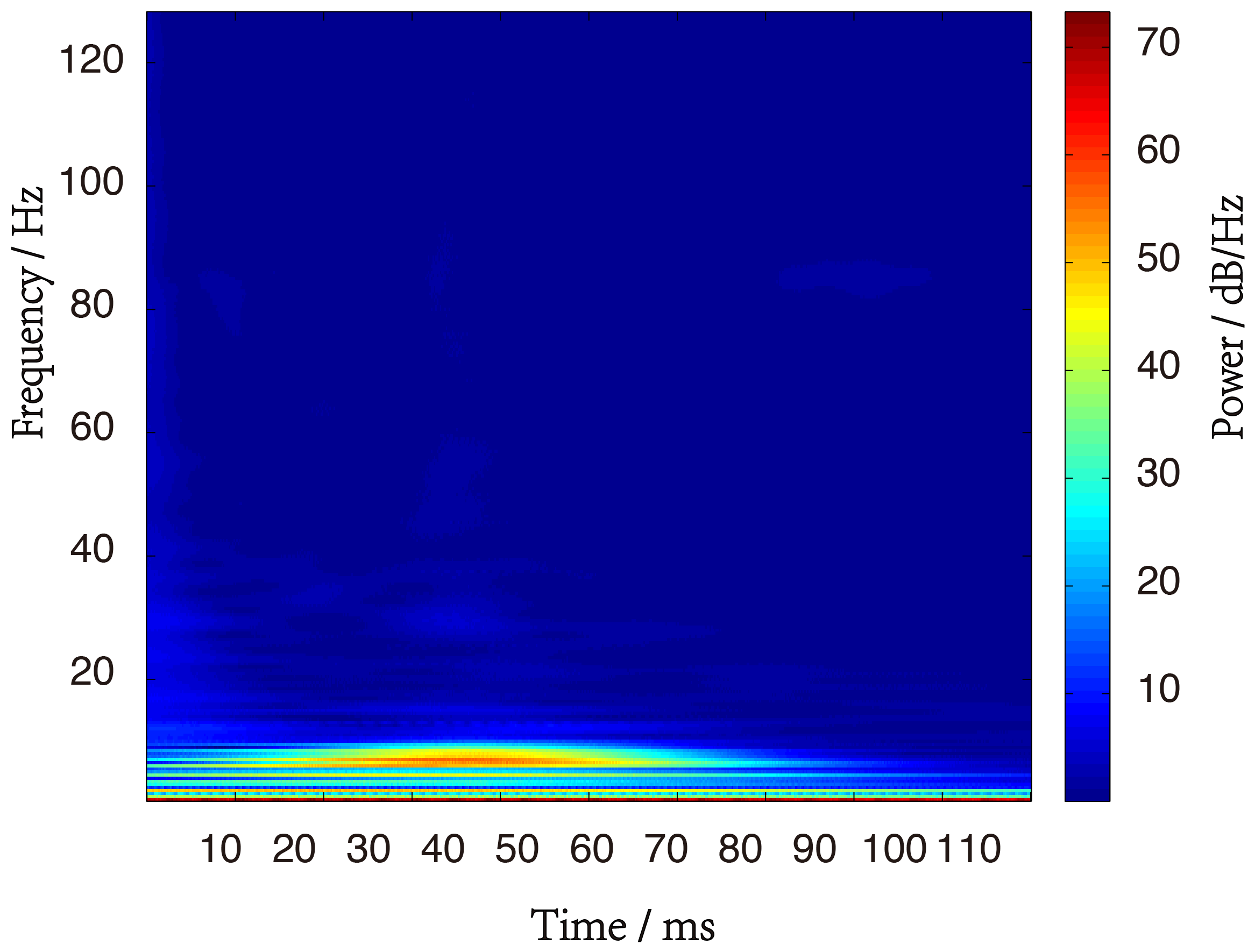}}
	\hspace{0.01in}
	\subfigure[]{
		\includegraphics[height=3.3cm,width=4cm]{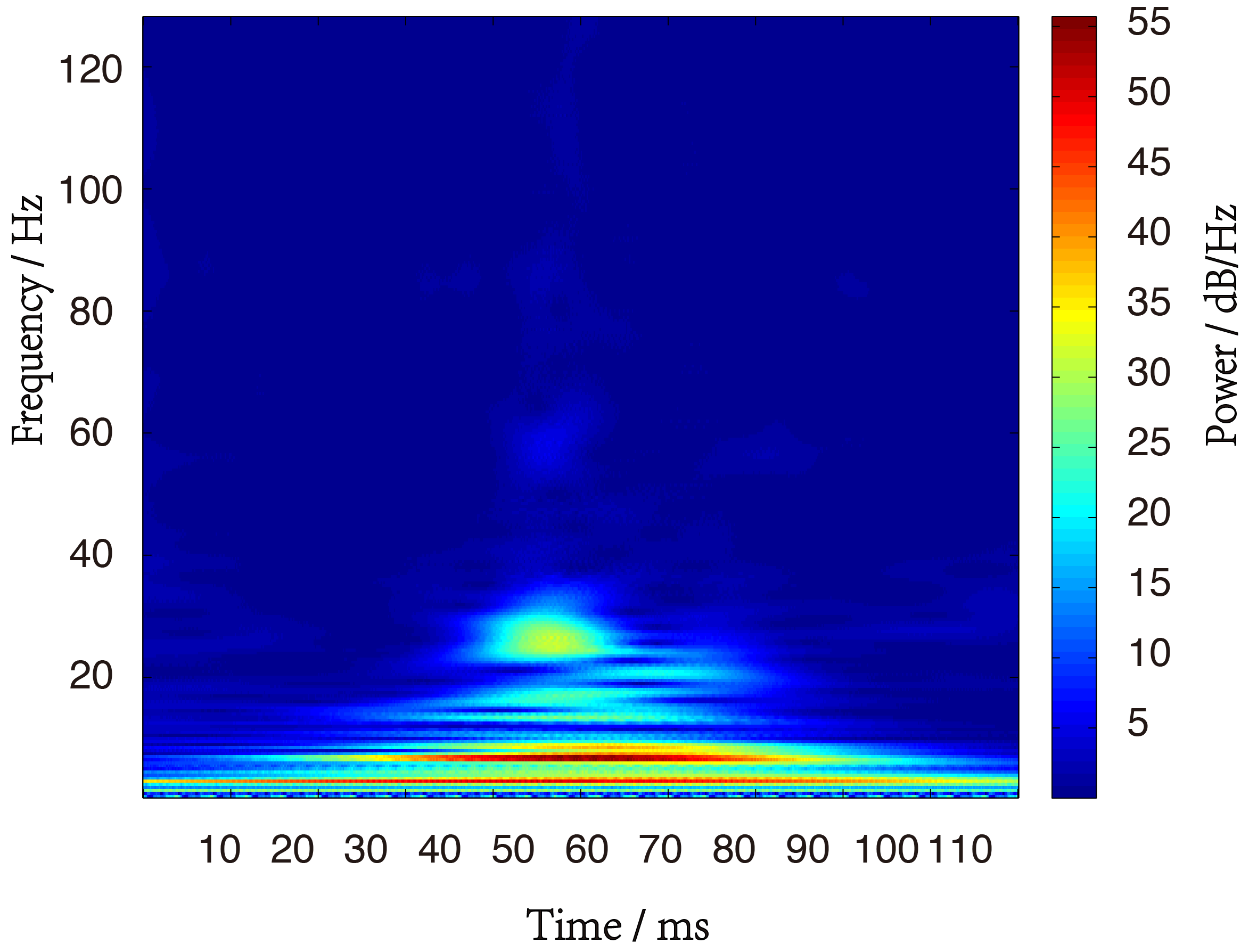}}	
	\caption{Time-frequency analysis using CWT. (a) Circle  (b) Square (c) Tick (d) Cross}
	\label{CWT}		
\end{figure}

Continuous wavelet transform (CWT) is an alternative way to see time and frequency change way of transformation. The difference between the CWT and the STFT is that the continuous wave of formation from the likelihood can be quite different and the result sometimes shows the quality of signal that cannot be used in classification. By applying below calculations to time domain signals, you can get scalograms like Fig. \ref{CWT}. Scalograms are just like spectrogram in STFT. In practice, we will first convert the value into frequency, so that the scalogram can be converted into wavelet time-frequency distribution.

\begin{align}
& CWT(t,\omega)=(\frac{\omega}{\omega_0})^{1/2}\int s(t')\varPsi^*(\frac{\omega}{\omega_0}(t'-t))dt, \label{equ3}\\
& CWT(t,\omega)=(\frac{(\omega_0/\omega)^{1/2}}{2\pi})\int S(\omega')\Psi^*(\frac{\omega_0}{\omega}\omega')e^{j\omega't}d\omega'. \label{equ4}
\end{align}

Fig. \ref{CWT} are four time-frequency distributions after processing CWT. As we can see, the difference is quite obvious. The frequency of circle gesture is always under 20Hz. And the square gesture is nearly 40Hz. As for the tick gesture, it has much lower frequency distribution. And cross gesture shows a gradual peak of frequency at the middle of the time window.

Though, there is enough difference between four gestures, plenty of differences between the result of time-frequency analysis cannot be noticed by eyes. So all these difference from the time-frequency analysis above then will be used in the classification phase in our experiments.

\subsection{Classification}

Using those informations we get the time-frequency analysis, we could use the convolutional neural network to do the classification.

Convolutional neural network is a classic classification method used anywhere. It is inquired from the idea that neuron connection from the animals brain. And how those signal transmit in animal brain neurons. They are typically used for detection and recognition of faces\cite{lecun2010convolutional}, texts\cite{lecun1998gradient} and logos, with great accuracy and robust performance. Also, CNNs were used to detect obstacles for vision-based android\cite{fan2010human} and segmentation of biological images\cite{lecun2004learning}. 

\begin{figure*}
	\centering{
		\includegraphics[height=5.5cm]{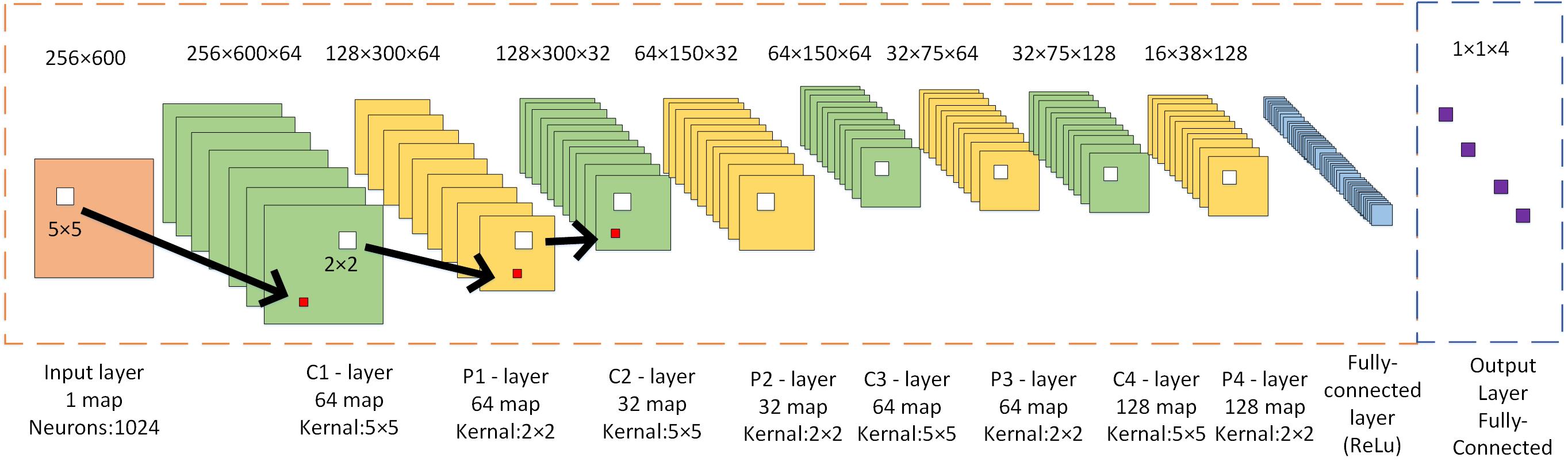}}
	\caption{CNN architecture with 10 layers}
	\label{CNN}		
\end{figure*}

Convolutional neural network usually consists of three typical layers stacked together: 1) Convolutional layer, 2) Pooling layer, 3) Fully connected layer. Our convolutional neural network consists of 10 hidden layers shown in \ref{CNN}, where we use four convolutional layers, four pooling layers and two fully-connected layers. Note that C-layer represents convolutional layer and P-layer represents pooling layer.

\textbf{Convolutional layer}: The first and the most important layer of the CNN is the convolutional layer. The layer's parameter consists of a series of core filters or kernels, which accord to certain activation functions. And their dimension can be defined by the input. In the forward passing of CNN, each kernel is sliding on the surface of the input matrix and go down if there is more dimensions. Then they compute the dot product accordingly of the kernel and the covering part of the matrix. Then it products the activation map of that kernel(filter). Intuitively, the CNN gradually learn the parameter of filter that activates when it detects some features at some position of the input matrix.

Then the C-layer will output the activation patterns which combines all the activation kernels of the whole input picture or signals. We can understand the kernel as the neuron of animals' brain that when they see some features in the new picture that accords to some feature of a seen animal such as cat, this neuron will light up.
The reason the area of each neuron will look is so small is that it have to be specific. Luckily, all kernel share the parameter in the same activation layer.

In our experiments, we adopted Rectiﬁed Linear Units(ReLU) as our activation function. The traditional activation function to model the output $f$ of a neuron regards to its input $x$ is Tanh-Sigmoid $f(x) = tanh(x)$ or Logistic-Sigmoid $f(x) = (1+e^{-x})^{-1}$. However, the saturating non-linearities are much slower. In 2012, Krizhevsky et al. proposed using neurons with activation function Rectiﬁed Linear Units(ReLU)\cite{krizhevsky2012imagenet}. And the results shows that training time required to reach 25\% of  error on the standard CIFAR-10 dataset for a just 4-layer CNN is much smaller than traditional sigmoid functions. Since our approach is based on the testing of big data, shorter training time is critical to our experiments.

\begin{figure}[htb]
	\centering{
		\includegraphics[height=4.5cm,width=6cm]{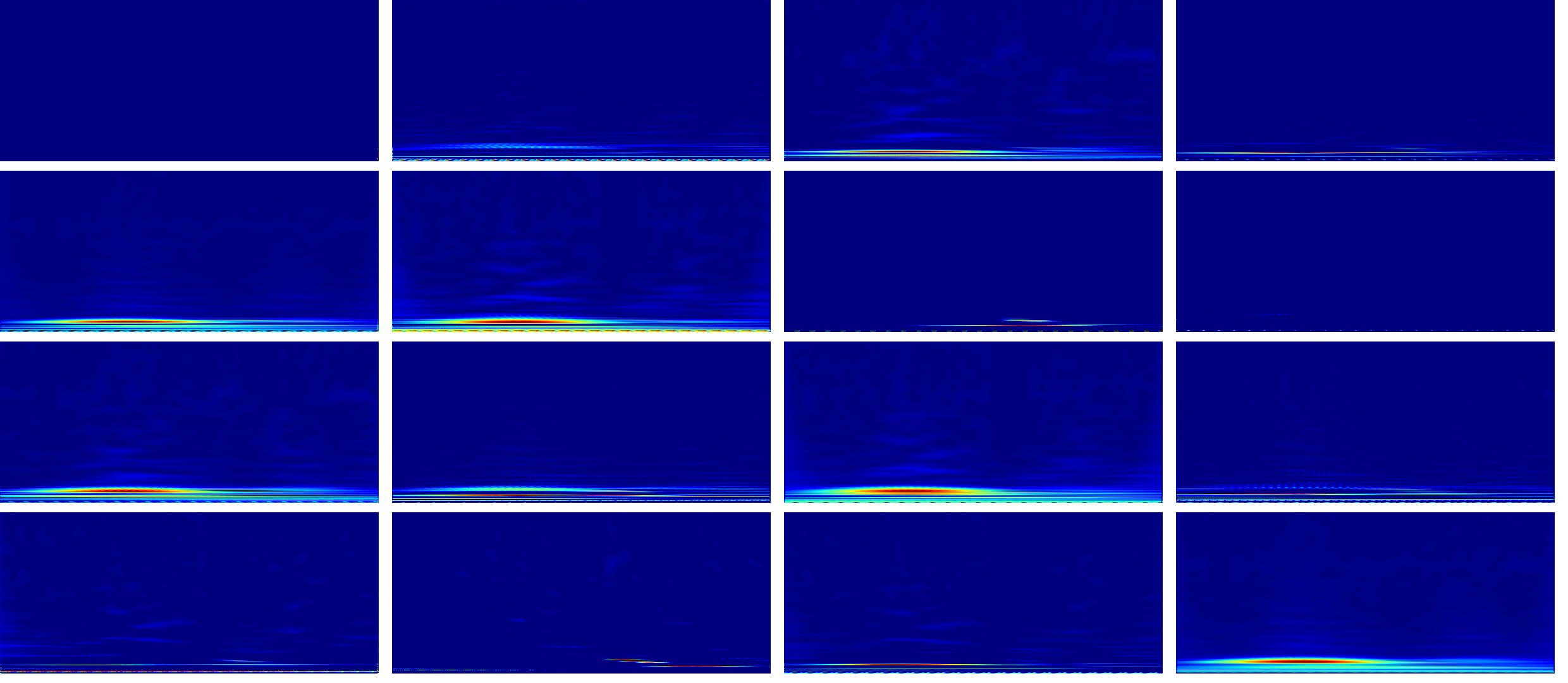}}
	\caption{The first 16 among 64 image activations of C1-layer in Fig. \ref{CNN} using 64 maps}
	\label{C1}		
\end{figure}

In our experiments, we use 64 maps of equal size (256$\times$600) in our first convolutional layer (C1-layer). A kernel of size (5$\times$5) (as shown in Fig. \ref{CNN}) is shifted over the valid region of the input matrix. All maps in one layer are connected to all the maps in the upper layer. And all the neurons of the same map share the same parametrization. Thus, the C-layer will output the activation patterns which combines all the activation kernels of the whole input picture or signals. The first 16 of total 64 activation maps of the C1-layer is presented in Fig. \ref{C1}. The parameterization of other two layers can refer to Fig. \ref{CNN}.

\textbf{Overlapping Pooling layer}: Another important layer of CNN is pooling layer, which conducts as non-linear down-sampling. Usually pooling layer is inserted between two convolutional layers. The pool layer is used to gradually do one simple thing, that is to do the down-sampling to the original size of kernels. Thus it can benefit the cost of complexity by reducing the calculation circles and the amount of parameters in the network.

In our experiments, to reduce the effect of over-fitting, we adopt overlapping pooling\cite{krizhevsky2012imagenet}. A pooling layer is constructed with a certain size of pooling neurons lined side by side and the distance between is $s$ units. Each unit calculates the maximum value, if this is a max-pooling way, of a selected area $p \times p$ located at the middle point of the pooling neuron. Because relative to other features, the real location of this feature is less we care so we can replace it with its rough location. When $s = p$, that is the commonly used pooling, which is quite traditional But if we have $s < p$, we can have overlapping pooling\cite{ciresan2011flexible}. That in our network, we set with $s = 2$ and $p = 3$, which is a typical value taking. The application of overlapping pooling reduces the error rate by 0.5\%. We notice that architecture with overlapping pooling are less easier to overﬁt during training.

\begin{figure}[htb]
	\centering{
		\includegraphics[height=4.5cm,width=6cm]{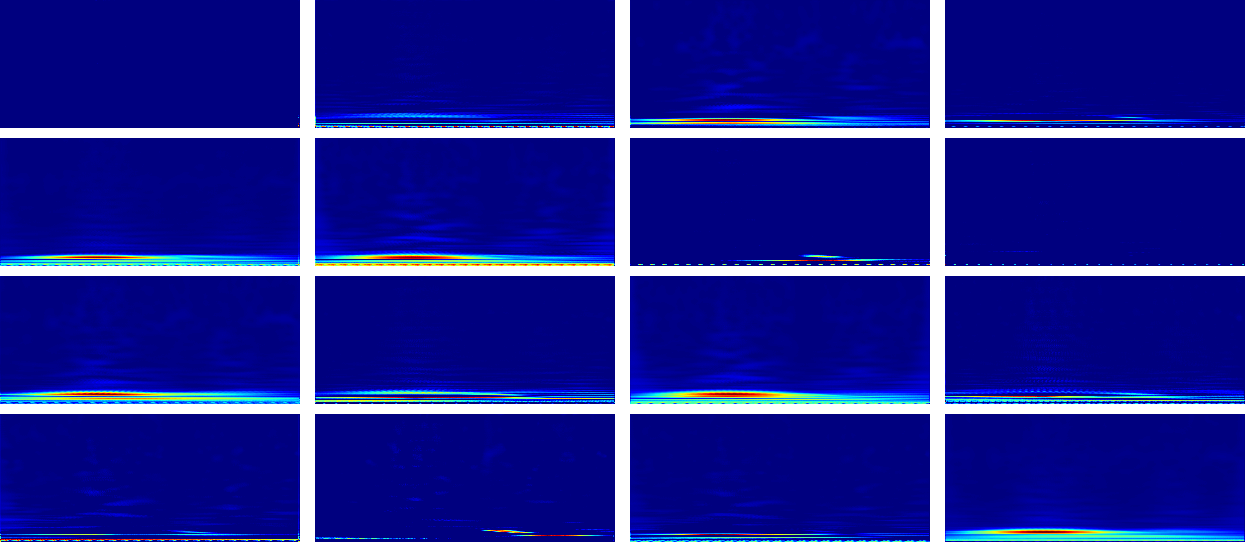}}
	\caption{The first 16 among 64 image activations of P1-layer in Fig. \ref{CNN} using 64 maps}
	\label{P1}		
\end{figure}

The first 16 of total 64  activation maps of the P1-layer is presented in Fig. \ref{P1}. Each block represents whether certain feature is activated in certain spatial of the input. The rest parameterization of other two layers please refer to Fig. \ref{CNN}.

\textbf{Fully connected layer}: After stacking of several convolutional and pooling layers, a fully connected layer will be placed as the end of the network and distribute the possibility of different classes.

\subsection{Details of training}

Besides, in our model, we adopt SGD with momentary of 0.8 and decaying weight rate of 0.001 in the training process\cite{krizhevsky2012imagenet}.  Choosing a less amount of decaying weight rate helps improving the effectiveness of model learning. Specifically, a less decaying weight rate can regulate the model training, while reducing the model’s training error in case of large database. The rule of updating weight $w$ is:

\begin{align}
& v_{i+1}=0.9\cdot v_i-0.0005\cdot\epsilon\cdot w_i-\epsilon\cdot \langle\frac{\partial L}{\partial w}|_{w_i}\rangle|_{D_i} \label{equ5}\\
& w_{i+1} = w_i+v_{i+1} \label{equ6}
\end{align}

in which $i$ is the iteration index and $v$ is the momentary variable. $\epsilon$ is the learning rate, and $\langle\frac{\partial L}{\partial w}|_{w_i}\rangle|_{D_i}$ is 
the average value over the $ith$ mini batch $D_i$ of the derivative value of the objective according to $w$, valued at the weight point $w_i$.

Initializations are quite important necessary to the speed of training. Because it can reduce the difficulty of the first amount of circles of learning. It gives positive inputs to the ReLU activation layers so the deal is made. We can initialize same amount of the weights in each layer with a Gaussian distribution of 0-mean and standard deviation 0.005. Besides, we use a constant value one to initialize the neural biases in the first , third convolutional layers, and the fully-connected hidden layers. The neuron biases in the remaining layers are initialized with the constant 0.

We use the same $\epsilon$ in all layers, and we fine-tune the parameters during the training. The fine-tuning rule about the rate is that we choose to give a 10 per cent of current rate when the training suddenly stopped or have little improvement\cite{krizhevsky2012imagenet}. And the initial learning rate is set as 0.01. And during the real training, it reduced three times and stopped at 0.00001 before termination. 

\subsection{Discussion}

The acquired 9600 samples are split into two sections. the first is training, and the second is testing. The first has 80 per cent of the samples and the second has 20 per cent.   Training ends as soon as the validation error comes to 0. Initialization are defines in the previous section that weights comes from a uniform random distribution with the range of (-0.05, 0.05). We trained the network for 50 cycles given 9600 samples, which took three to four hours on twenty Intel Xeon 2.4GHz CPUs.

We select the trained network with the lowest validation error. Then we test it on the remaining samples. The result shows that the accuracy of the recognition rate is nearly 98\%.

\begin{figure}[tb]
	\centering
	\subfigure[]{
		\includegraphics[height=3.5cm]{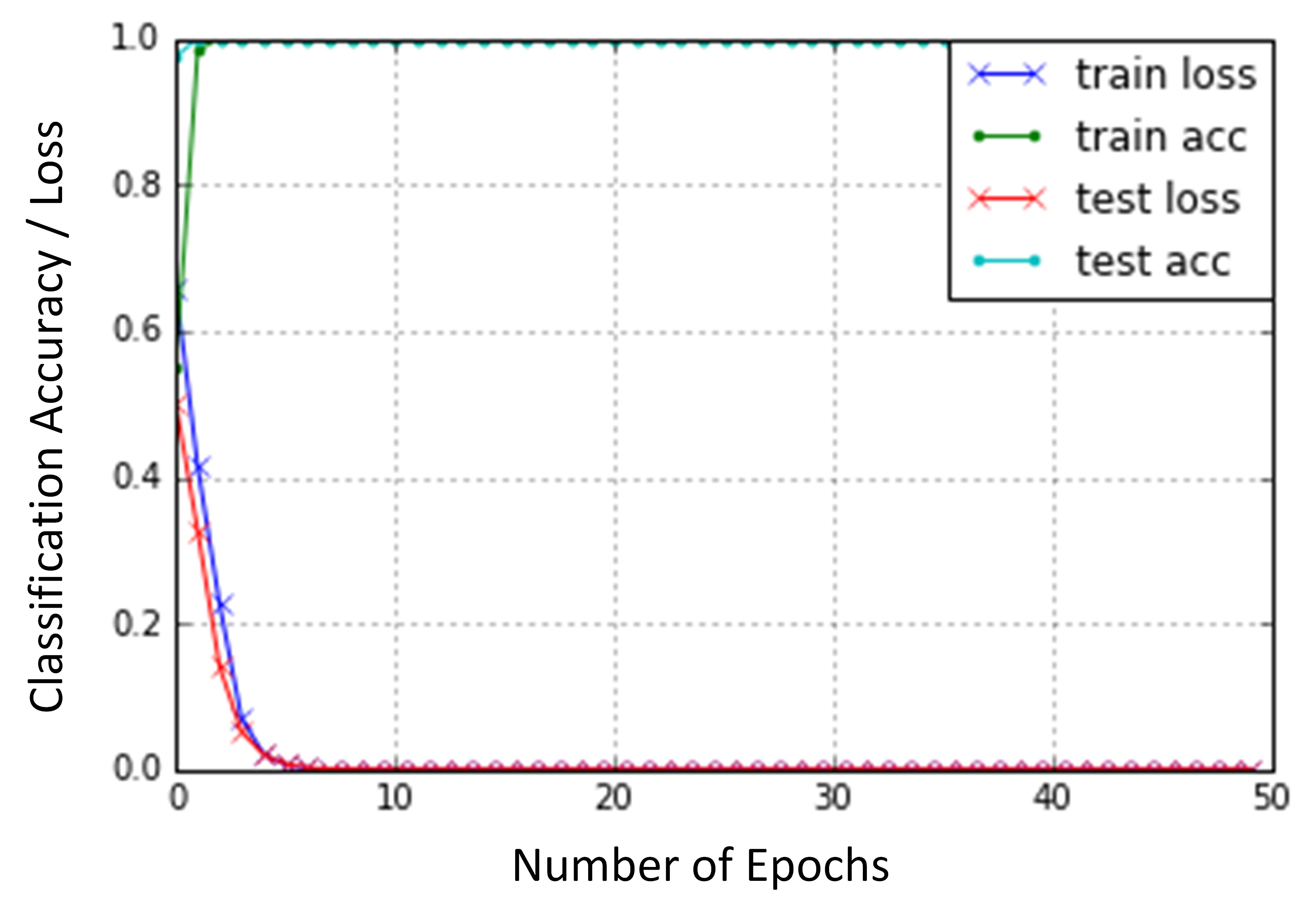}}
	\hspace{0.1in}
	\subfigure[]{
		\includegraphics[height=3.5cm]{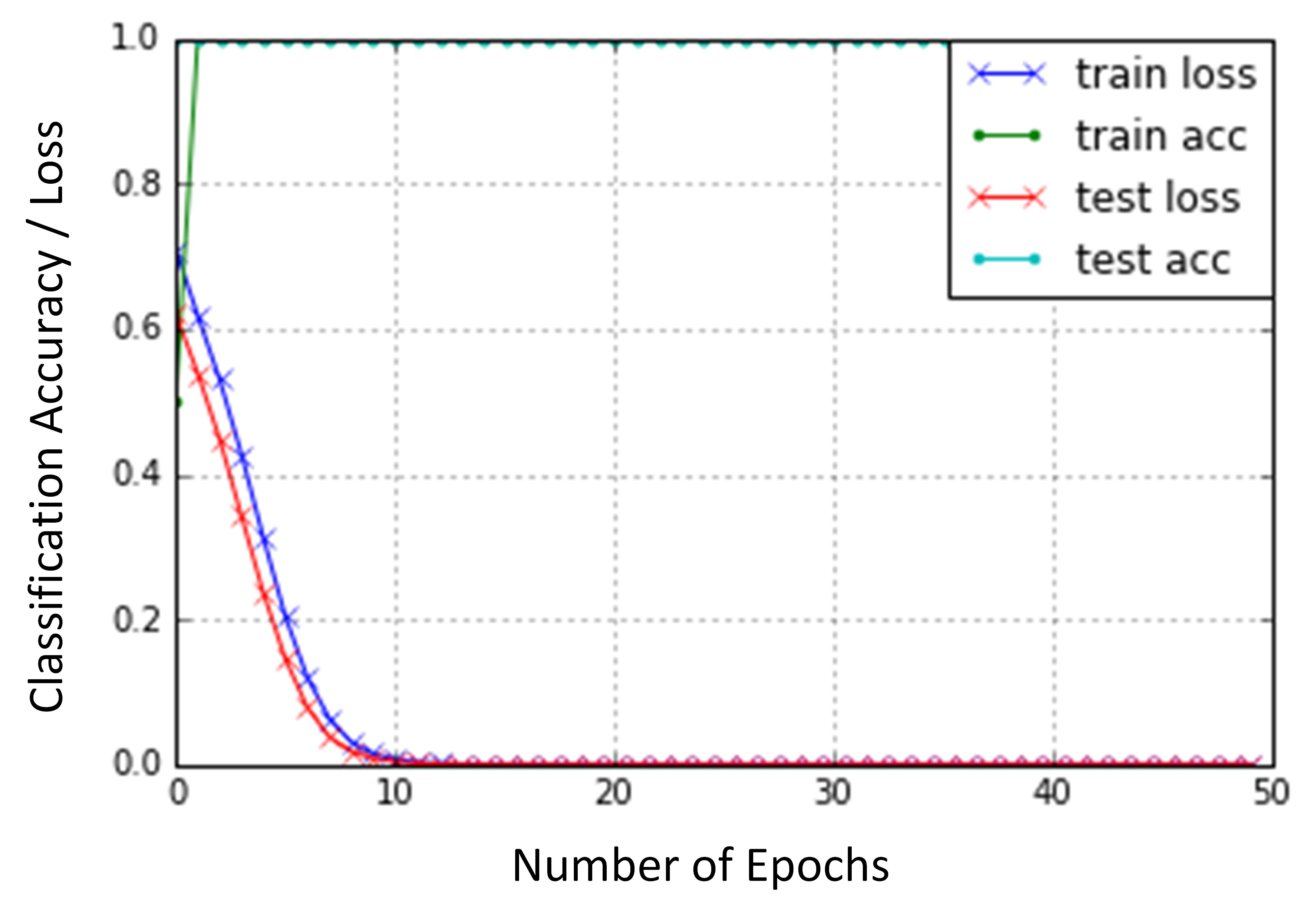}}	
	\hspace{0.1in}
	\subfigure[]{
		\includegraphics[height=3.5cm]{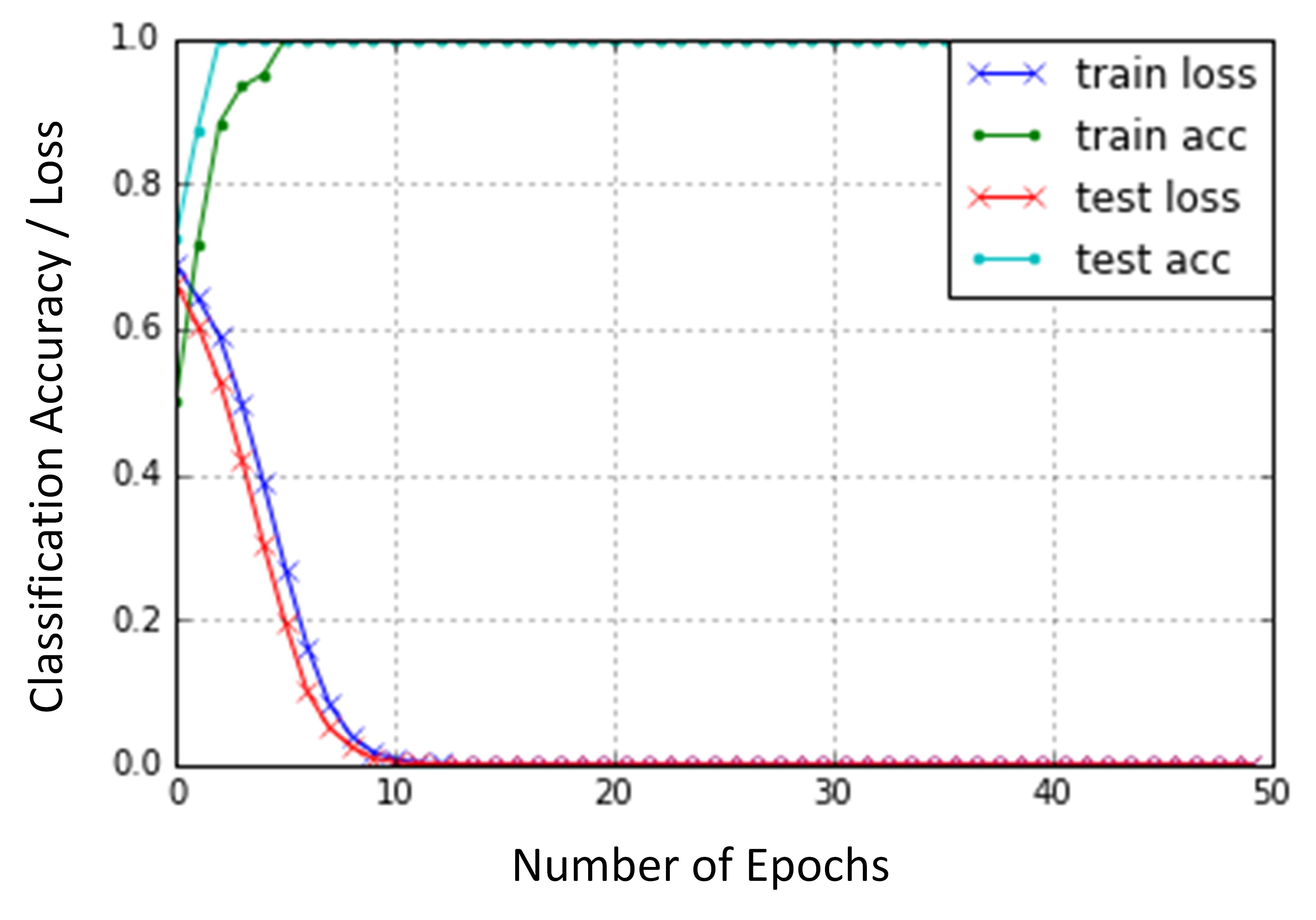}}	
	\caption{Classification accuracy and loss for training and testing datasets captured in different distances: (a)$d=0.1$; (b)$d=0.2$; (c) $d=0.5$ (in meters) using CNN architecture shown in Fig. \ref{CNN}. Results are averaged using 50 training and testing sets, where each set is selected randomly.}
	\label{dis}
\end{figure}

In order to evaluate the effect of distance and scale of gesture on the accuracy of recognition, we test the CNN with samples captured in different distances and give the classification results in Fig. \ref{dis}. In the figures, the accuracy and loss of training and testing are all plotted Note that the data loss is defined as an average over the data losses for every individual example. That is $L=\frac{1}{N}\sum_i{L_i}$ where $N$ is the number of training data. As we can see that, the epochs to reach the highest accuracy recognition rate is increasing. When the distance between hand gesture and radar sensor is 0.1 meter, it only takes 5 epochs to finish training and achieve the highest accuracy. And when the distance increases to 0.5 meter, it takes more than 10 epochs to reach the lowest loss rate. The result means that with the distances increasing, the difficulties to find a suitable classifier also increase. 

\begin{figure}[tb]
	\centering
	\subfigure[]{
		\includegraphics[height=3.5cm]{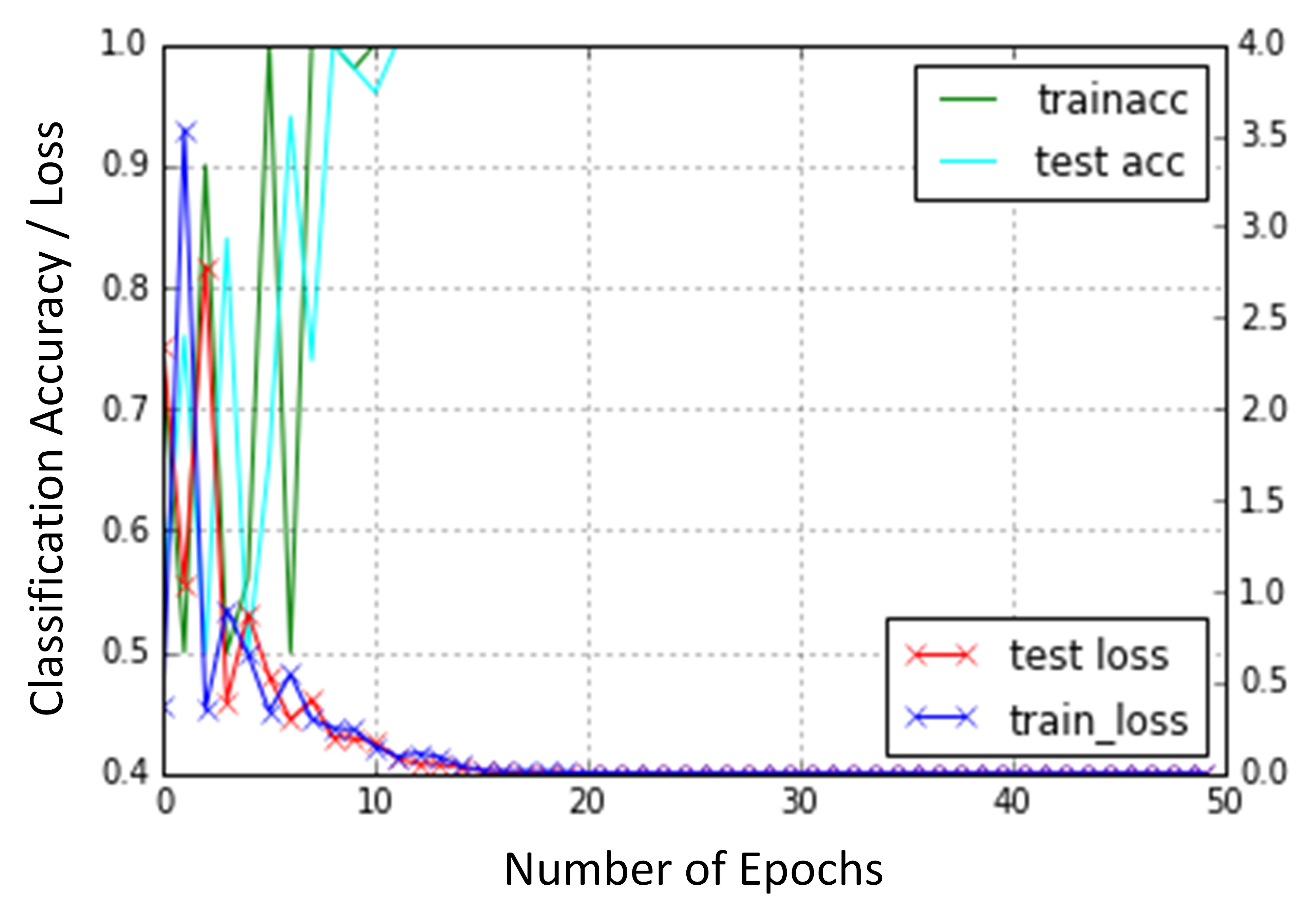}}
	\hspace{0.1in}
	\subfigure[]{
		\includegraphics[height=3.5cm]{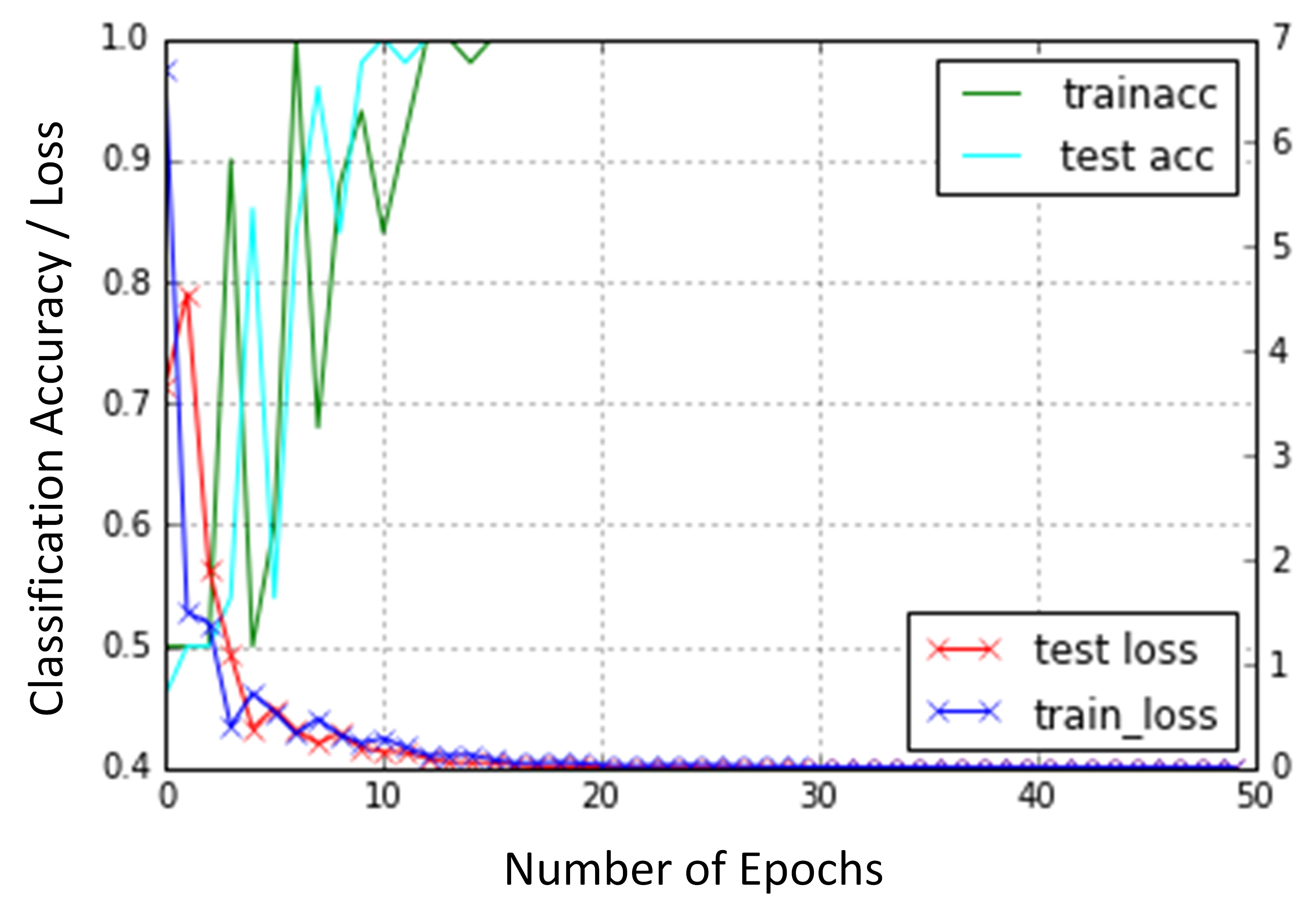}}	
	\caption{Classification accuracy and loss for training and testing datasets captured with different scale of gesture: (a) $r=0.2$; (b) $r=0.5$ (in meters); using CNN architecture shown in Fig. \ref{CNN}. Results are averaged using 50 training and testing sets, where each set is selected randomly.}
	\label{scale}
\end{figure}

Besides, we investigate how the scale of gesture affects the accuracy of recognition. The result is shown in Fig. \ref{scale} with the scale of certain hand gesture larger, the loss rate of recognition does not change much. From the perspective of accuracy , the loss is apparently bigger when the scale of hand gesture became larger.

\section{Conclusion}
This paper proposes a Doppler-Radar based hand gesture recognition system using convolutional neural networks. Our system adopted Doppler-radar sensor with dual receiving channels at 5.8 GHz to acquire large hand gesture database of four standard hand gestures. We applied two time-frequency analysis, short-time Fourier transform and continuous wavelet transform, to receive hand gesture signal. And then we use convolutional neuron networks as classifier of four different gestures based on the results of time-frequency.  Experiments achieve gesture recognition performance with a classification rate of 98\%. 

Besides, related factors such as distance and scale of hand gesture are investigated. Experiments shows that with the distance increasing, the convolutional neural network needs more epochs to reach the finest accuracy recognition rate. So does the impacts of gesture scale over the loss rate of recognition. In the future study, our group will focus on more impacting factors like how different hand of different people and the speed of the gesture affect the recognition accuracy.

To sum up, our proposed system proves the feasibility of using narrow-band microwave radar of 5.8 GHz in hand gesture recognition and shows potential of wider application of microwave radar in the pattern recognition field.\cite{goodfellow2014generative}\cite{radford2015unsupervised}\cite{dai2017deformable}

\section*{Acknowledgments}
This work is supported by Intel under agreement No. CG\# 30397855.

\bibliographystyle{IEEEtran}
\bibliography{5G}

\end{document}